\documentclass{article}
\usepackage[preprint]{neurips_2023}

\usepackage[utf8]{inputenc} 
\usepackage[T1]{fontenc}    
\usepackage{xcolor}         
\usepackage[colorlinks=true,citecolor={magenta},linkbordercolor={white}]{hyperref}       
\usepackage{url}            
\usepackage{booktabs}       
\usepackage{amsfonts}       

\usepackage{nicefrac}       
\usepackage{microtype}      

\usepackage{graphicx}
\usepackage{subcaption}
\usepackage{caption}
\usepackage{float}
\usepackage{amsmath}
\usepackage{amssymb}
\usepackage{geometry}
\usepackage{hyperref}

\title{Why does in-context learning fail sometimes? Evaluating in-context learning on open and closed questions.}

\author{%
    Xiang Li$^{1, 2}$ \quad
    Haoran Tang$^{2}$ \quad
    Siyu Chen$^2$ \quad
    Ziwei Wang$^2$ \quad
    Ryan Chen$^2$ \quad
    Marcin Abram$^{1, 3}$\\[4pt]
    $^1$Department of Physics and Astronomy, University of Southern California, Los Angeles, CA, USA\\[2pt]
    $^2$Department of Computer Science, University of Southern California, Los Angeles, CA, USA\\[2pt]
    $^3$Information Sciences Institute, University of Southern California, Los Angeles, CA, USA\\[4pt]
    \texttt{\{limike,haoranta,schen809,zwang476,rchen259,mjabram\}@usc.edu}\\
}

\begin{document}

\maketitle

\begin{abstract}
    We measure the performance of in-context learning as a function of task novelty and difficulty for open and closed questions. For that purpose, we created a novel benchmark consisting of hard scientific questions, each paired with a context of various relevancy. We show that counter-intuitively, a context that is more aligned with the topic does not always help more than a less relevant context. This effect is especially visible for open questions and questions of high difficulty or novelty. This result reveals a fundamental difference between the treatment of close-form and open-form questions by large-language models and shows a need for a more robust evaluation of in-context learning on the variety of different types of questions. It also poses a new question of how to optimally select a context for large language models, especially in the context of Retrieval Augmented Generation (RAG) systems. Our results suggest that the answer to this question can be highly application-dependent and might be contingent on factors including the format of the question, the perceived difficulty level of the questions, and the novelty or popularity of the information we seek.
\end{abstract}

\section{Introduction}

    Despite their indisputable successes \citep{Bommasani2021, Drori_2022, Chang2024}, Large Language Models (LLMs) often struggle to answer challenging questions \citep{rawte2023troubling}. While they can achieve superhuman accuracy on many benchmarks \citep{Luo2024}, they also suffer from hallucinations \citep{Ye2023, Azamfirei2023}, lack of coherence \citep{Xie2023Qiming}, and are prone to cognitive errors \citep{Jones2022, Hagendorff2023}.

    To make the difficult situation even worse, it is not always easy to detect mistakes committed by LLMs since their responses are often presented in a way that emulates correct and coherent answers \citep{Bender2021, Scheurer2023}.
    For practical reasons, many existing benchmarks only test the ability to answer either closed \citep{Chang2024} or easy-to-verify questions, e.g., regarding common knowledge \citep{bisk2019piqa, clark2018think} or questions that can be algorithmically verified \citep{Srivastava2024}.

    Another challenge concerns domain generalization and domain shift problems, resulting in the need to constantly update your machine learning models to account for the evolution of various trends in your data \citep{Zhou2022}. However, improving the performance of pre-trained LLMs for specific tasks by fine-tuning is both expensive \citep{Bender2021, Luccioni2023} and technically challenging \citep{pmlr-v202-kandpal23a, Gaspers2022}. While some techniques like Low-Rank Adaptation (LoRa) can reduce the cost of training \citep{Hu2011}, it does not solve the main issue, namely, how to allow LLMs to leverage new pieces of information that were not a part of the initial training corpus \citep{Liu2017}.

    One approach to the issue might be in-context learning \citep{brown2020language}, where LLMs effectively learn to solve a given problem leveraging a limited number of examples without updating the model parameters. Namely, in-context learning incorporates question-solution pairs in the input prompt, allowing LLMs to detect the logic and patterns of those examples, subsequently improving the LLMs output accuracy. It enables LLMs to acquire new knowledge in the inference time and utilize it in subsequent responses. This technique significantly reduces the complexity of improving the LLMs performance compared to alternative approaches such as fine-tuning \citep{min2022rethinking}. 
    It should also be noted that the effectiveness of the popular Retrieval-Augmented Generation (RAG) techniques relies heavily on the strength of in-context learning    \citep{gao2024retrievalaugmentedgenerationlargelanguage}, as discussed later.

    In this paper, we focused on the question of how various types of context improve the effectiveness of in-context learning when answering challenging questions. We noticed a surprising behavior. Namely, depending on the difficulty and novelty of the question, and depending on the fact whether the question is of the open or closed type, the relation of the measured performance of the model to both the perceived and quantified relevancy of the context \emph{varies}. Notably, the measured in-context learning performance of GPT-4 was positively correlated to context relevancy in two benchmarks with closed-form questions but negatively correlated in our benchmark with open-form questions, indicating different utilization of context depending on the form of the received questions.

    In the next sections, we introduce our novel dataset, which comprises 160 unique question-response pairs from the fields of physics and computer science with varying levels of difficulty. For the purpose of evaluation, each question is accompanied by one of four types of context (including \emph{no context} to serve as a control group) and paired with a generated answer from GPT-4. In the subsequent sections, we detail our grading scheme and present the results aggregated from each of our graders. Next, we compare our findings with the existing work by \citet{min2022rethinking}, highlighting a notable discrepancy in the measured effectiveness of the context. To elucidate this difference, we delve deeper into the nature of the problem, discovering that the main impact comes from the open or closed form of the questions, with additional effects related to the difficulty or novelty of those queries.
    To further strengthen our analyses, we then compare the performance improvement associated with in-context learning across a range of context relevancy using two additional close-ended question datasets, MetaICL \citep{min2022metaicl} and NephSAP \citep{wu2023openicl} and we contrast the results with our findings harvested with the help of our open-ended question dataset. Following this, in the discussion section, we discuss the impact of our work, especially in the context of the RAG systems, future research directions, and other methods that enhance LLM performance 
    
\section{Related Work}

    \paragraph{Large Language Models.}
    LLMs have shown remarkable capabilities in various tasks, including code generation \citep{kojima2023large, siddiq2023generate}, text summarization \citep{sahu2023enchancing}, and database query optimization \citep{Li2023}. They demonstrate a surprising ability to perform in-context learning, where an LLM ``learns'' to do a task simply by conditioning on a prompt containing input-output examples, achieving state-of-the-art (SOTA) results on various benchmarks. However, there has been little understanding of how the model leverages the context and what makes in-context learning work.
    In addition, their performance significantly depends on the contextual information provided and, as discussed in this paper, on the form and type of the queries.

    \paragraph{In-Context Learning.} In-context learning has been a focal point in recent research. Unlike tra\-ditional fine-tuning methods, in-context learning adapts models to unseen tasks by incorporating examples directly into the input context, as highlighted by \citet{brown2020language}. \citet{xie2022explanation} discussed how in-context learning can be understood as implicit Bayesian inference, where models infer latent concepts to generate coherent responses. Techniques such as chain-of-thought prompting \citep{wei2023chainofthought, press2023measuring, wang2022iteratively, zhou2023leasttomost, imani2023mathprompter, besta2023graph} have shown significant improvements in reasoning tasks. Recent frameworks like OpenICL \citep{wu2023openicl} have further streamlined the implementation of in-context learning by providing unified and flexible tools for integrating various retrieval and inference methods.

    Many recent research focuses on the example selection strategies of in-context learning. One of the most common strategies is to select examples for demonstration based on similarity in the embedding space \citep{liu2022what, qin2023in, gao2021making}. In-context learning seems robust to label-noise, as indicated by work of \citet{min2022rethinking}, in which authors show that demonstrations, even one with randomly shuffled labels, can still significantly improve LLM's performance in the MetaICL dataset. 

    \paragraph{Evaluation Benchmarks.} Benchmarking is essential for understanding LLM performance across different domains. Existing benchmarks like AGIEval \citep{zhong2023agieval}, ChenLLMBench \citep{guo2023chemllmbench}, SCIEval \citep{sun2023scieval}, PIXIU \citep{xie2023pixiu}, and MME \citep{fu2024mme} provide comprehensive datasets for evaluating LLMs. While these benchmarks are useful for understanding the general capabilities of LLMs, they do not capture the complexity of more open-ended and context-sensitive queries. Here, the added value of our work, as we believe the novel open-question validation set we created, fills that gap.

\section{Originality and general impact of the work assessment}

    \paragraph{Originality.}
    In this paper, we argue that closed questions, such as multiple-choice or fill-in-the-blank formats, do not adequately reflect the challenges posed by open questions that require deep understanding and synthesis of information from diverse contexts. While \citet{min2022rethinking} have shown that context significantly affects LLM performance, they have not quantified how different levels of context relevancy impact responses to different types of questions. Our research addresses this gap by creating a novel benchmark that focuses on open, challenging questions. These questions are paired with various types of contexts to systematically evaluate how context affects LLM performance. 

    \paragraph{Impact of the paper.}
    Furthermore, our work suggests areas for improving the performance of Retrieval-Augmented Generation (RAG). Current RAG studies focus on providing context during model inference. Given our observation of the inconsistent relationship between the relevance of context and model performance for different question types (open-form and closed-form), we believe that the context retrieved by comparing vector similarity using RAG may not always correlate with the most useful context for enhancing LLM inference performance and does not mitigate issues such as hallucinations and logic errors. We propose that the type of context selected should be tailored to the attributes of the type of questions with several practical propositions of the retrieval regions outlined in the discussion.

\section{Is more relevant context always better?}

\subsection{Novel question bank and evaluation methodology}

    To investigate the relationship between the relevance of context and the performance of large language models (LLMs), we created an open-form questions dataset comprising physics and computer science questions of varying difficulty levels and originality. Next, we prepared contexts with four different levels of relevancy for each question in our dataset.

    The selected questions cover the following areas: quantum mechanics, physics for life science, electromagnetism, classical mechanics, and computer science. Solutions usually involve a combination of complex calculations and the application of conceptual knowledge. Each question is categorized under one of the three different difficulty levels: easy, medium, and hard. The difficulty of the question is defined by the grader according to their perceived complexity of the question. Additionally, each question is also categorized under one of three originality categories: known, paraphrased, and original. Known questions can be found online or in textbooks, paraphrased questions are modified versions of known questions, and original questions were handcrafted by the authors of this paper. 
    
    For each question, we created a ground truth answer for scoring reference and four context types with different levels of relevance. The four context types are: (1) ``no context'' to serve as a control group, (2) ``irrelevant context'', which consists of text on topics that do not match the subject of the question, (3) ``vague context'', which incorporates some topics or keywords related to the question, and (4) ``relevant context'', which provides reasoning context for the question, or answer to a highly related question. Next, for each unique pair of question-context, we generated a response employing the OpenAI's\ \emph{gpt-4-1106-preview} model. 
    
    After retrieving the responses, we constructed 160 question-response pairs, each accompanied by the corresponding ground truth.
    %
    Aware that human grading can be subjective, we decided that each question would be evaluated by six independent graders using a pre-defined scoring sheet. This gave us 960 evaluation responses in total.

    The Supplementary Material includes examples of the questions and context types, as well as the evaluation sheets.    

\subsection{Evaluation}

    Our evaluation system comprised three main categories, 
    \emph{Completeness and Relevancy} (5 points),
    \emph{Logic and Reasoning} (5 points), and
    \emph{Truthfulness (understood as lack of hallucination)} (5 points).

    In addition, graders had the option to identify specific problems in the responses, such as \emph{hallucinations}, \emph{omission}, \emph{irrelevant}, \emph{calculation error}, and \emph{logic error}. They could also highlight portions of the responses as \emph{incorrect}, \emph{correct}, or \emph{irrelevant}. An open response section was provided for graders to give comments and feedback about the generated responses. Finally, graders were asked to rate their confidence in their own grading. These options allowed us to gain deeper insights into the grading process and to assess the quality of the generated responses in detail.
    A screenshot of the scoring interface can be found in the Supplementary Material.

    Each grader may have different biases and varying levels of expertise. To enhance the accuracy and reliability of our evaluation, we ensured that all graders assessed all 160 questions. This approach was essential for obtaining consistent and accurate results. By having multiple graders evaluate each response, we mitigated individual biases and ensured a more comprehensive assessment. This method captured a broader range of perspectives and expertise, leading to a more robust and reliable evaluation of the generated responses. As demonstrated later, this comprehensive grading significantly improved the accuracy and consistency of our findings.
    
\subsection{Results}


    \begin{figure}
        \centering
        \includegraphics[width=1\textwidth]{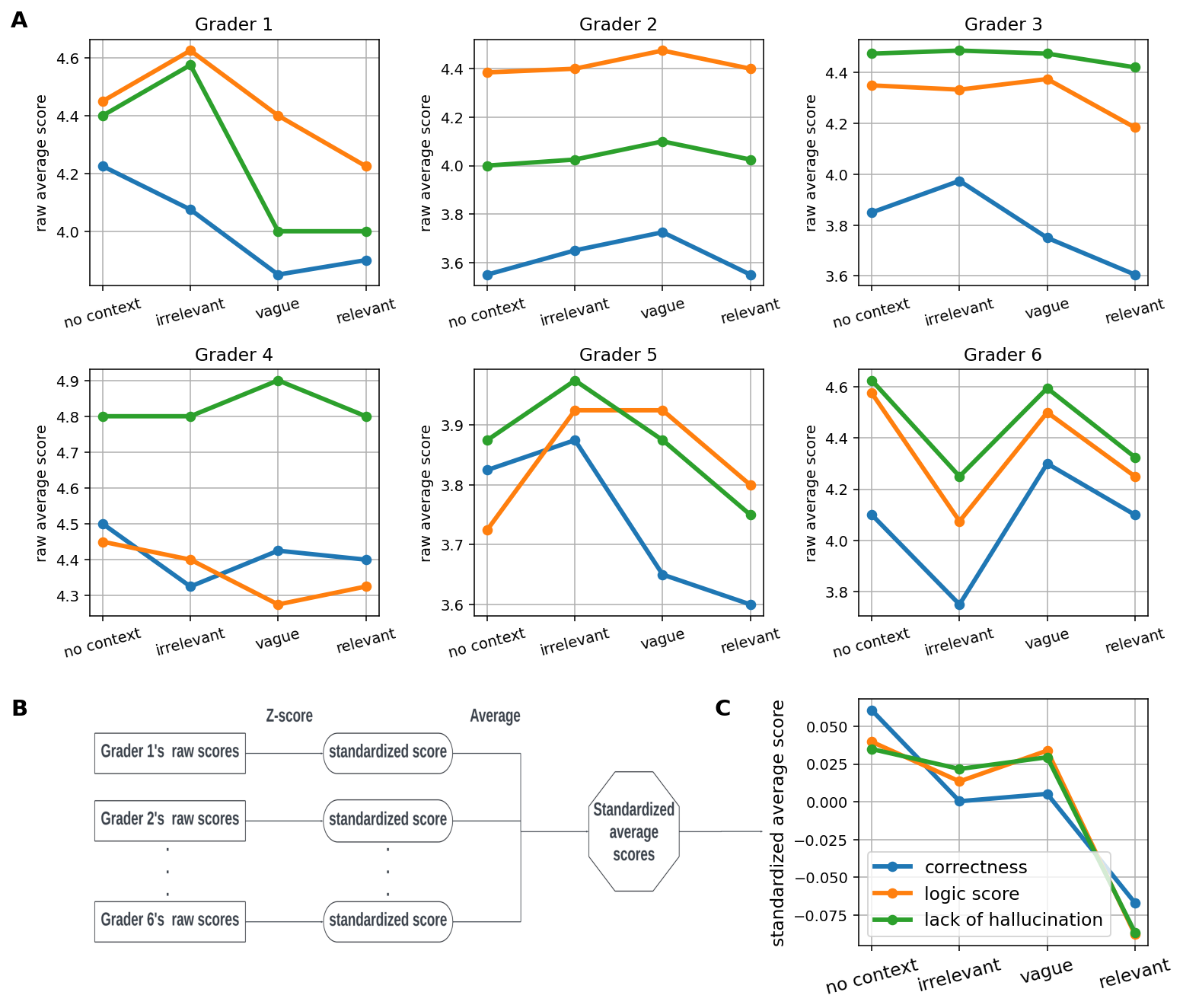}
        \caption{(A) Raw average scores of generated responses for each context type (\emph{no context}, \emph{irrelevant context}, \emph{vague context}, and \emph{relevant context}) evaluated for \emph{Completeness and Relevancy (Correctness)}, \emph{Logic and Reasoning (Logic Score)}, and \emph{Truthfulness (lack of hallucination)}, assessed by six different graders. (B) The process of standardizing raw scores from each grader to calculate the overall standardized average scores. The raw scores are converted to Z-scores, which are then averaged to obtain standardized average scores. (C) Standardized average scores of generated responses for each context type aggregated across all graders.} 
        \label{fig:main_results_plot}
    \end{figure}

\subsubsection{Context Relevance}

    To illustrate the correlations between the context types and the quality of the corresponding generated responses, in Fig. \ref{fig:main_results_plot} panel A, we show the raw average scores of each context type for each grader. Notably, the results are rather noisy, with each grader having an individual tolerance for different types of errors, resulting in different reference levels for each of them.

    By design, each question was evaluated by each grader. This additional redundancy allows us to standardize the scores for each grader and then average them, resulting in reduced variance in the final results. This aggregation procedure is depicted in Fig. \ref{fig:main_results_plot} panel B.
    
    

    As a result, although the raw scores displayed differences in trends and values across all three grading rubrics, a clear trend appeared after we applied the aggregation procedure, as depicted in Fig.~\ref{fig:main_results_plot} panel~C. Counter-intuitively, a higher standardized average score was associated with \emph{no context}, and the lowest score with the \emph{relevant} context.
    
\subsubsection{Difficulty Levels and Originality Types}

    \begin{figure}
        \centering
        \includegraphics[width=1\textwidth]{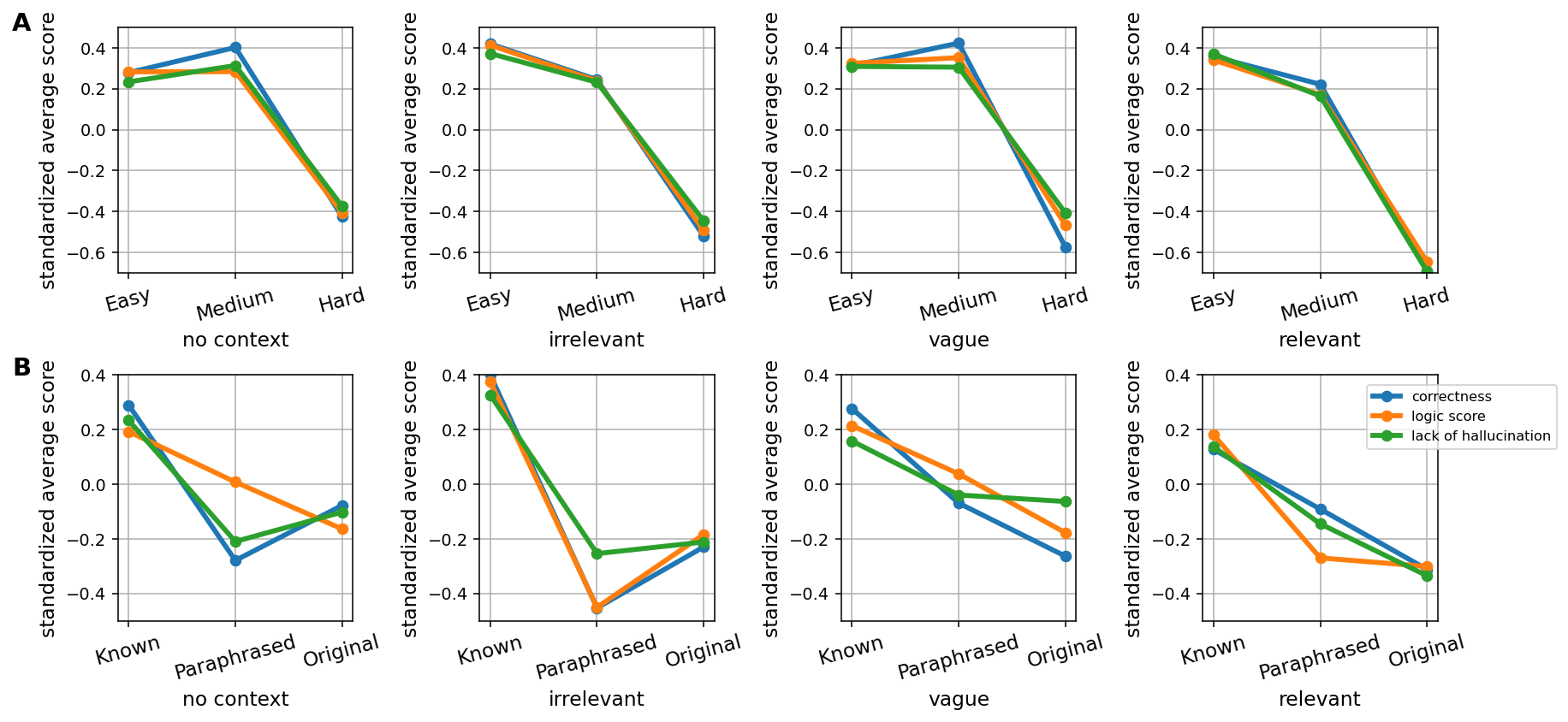}
        \caption{(A): Standardized average scores of generated responses for each context type (\emph{no context}, \emph{irrelevant context}, \emph{vague context}, and \emph{relevant context}), categorized by three levels of question difficulty (\emph{easy}, \emph{medium}, and \emph{hard}) for correctness, logic errors, and lack of hallucination. (B):~Standardized average scores of generated answers for each context type, subdivided into \emph{known}, \emph{paraphrased}, and \emph{original} categories, evaluated for correctness, logic score, and lack of hallucination.} 
        \label{fig:difficulty_originality}
    \end{figure}
    
    To investigate how the difficulty of questions affects the quality of generated responses, we compared the results across three difficulty levels (\emph{easy}, \emph{medium}, and \emph{hard}) for each of the four context types, as shown in Figure \ref{fig:difficulty_originality}, panel A. We can observe a clear trend of decreasing scores as the difficulty of the questions increased from medium to hard, indicating that GPT-4's performance declines with higher question difficulty.
    This also indicates that human-perceived difficulty of the question was in fact, correlated with the factual difficulty experienced by GPT-4, a result interesting on its own.
    For easy and medium-difficulty problems, GPT-4 generated responses with similar scores, indicating that the alignment between the human-perceived and machine-perceived difficulty has its own limits.
    

    In Figure \ref{fig:difficulty_originality}, panel B, we show the comparison between the aggregated standardized average score for the different levels of originality types for each context type. It is evident that GPT-4 scores highest for known questions, likely because these questions were part of its training data, and therefore GPT-4 has a higher chance to answer them correctly. Interestingly, the score for known questions given irrelevant context is twice as high as that for relevant context. This suggests that irrelevant context might be more helpful than relevant context for known questions, at least for the open type of question, as measured here.

\subsubsection{Result comparison}
    
    In this section, we combined the standardized scores from all graders and compared them across different context types. Our results indicate that, on average, the responses generated with no additional context or with the help of irrelevant context are of higher quality than the responses generated for queries incorporating highly relevant context.
    This result is in striking difference to results of \citet{min2022rethinking}. To further understand this discrepancy, in the next section, we replicate the key findings of \citet{min2022rethinking}, and we discuss what might cause the difference in the behavior.


\section{Citical comparison with existing study}

\subsection{Intro}

    \citet{min2022rethinking} demonstrates that in-context learning allows us to achieve significantly better results compared to the ``no context'' case. In addition, the authors show that in-context learning is robust to label noise. Namely, the authors show that context with randomly shuffled labels and ``golden'' context (with correct labels) have similar effects in enhancing the quality of generated responses for closed questions, such as multiple choice and true/false questions.

    However, to investigate the striking difference in the observed trends and to eliminate the effect of different versions of ChatGPT playing a potential role here, we decided to replicate the key results from \citet{min2022rethinking} using precisely the same framework as above and using the same version of the LLM, namely \emph{gpt-4-1106}.

    For the replication, we decided to use two different existing benchmarks, MetaICL \citep{min2022metaicl} and a dataset from NephSAP \citep{unknown}. The only significant element, differentiating this study from our previous evaluations, is that both of these datasets contain close-form questions.

\subsection{Data and Methodology}

    Our evaluation of in-context learning of closed-form questions involves two datasets. For the MetaICL dataset, we take a subset of 10 different tasks, each containing multiple-choice questions. For the NephSAP dataset, we take multiple-choice questions within 20 different subjects. Details about tasks, subjects, and sample questions can be found in the Supplementary Materials.

    We conduct an 80-20 train test split for both the MetalCL dataset and the NephSAP dataset. For each multiple-choice question in the test set, we generate a response using the \emph{gpt-4-1106-preview} model. We do it three times: once without any context, once with a randomly sampled demonstration with a different task or subject from the training set of the dataset, and once with a randomly sampled demonstration with the same subject or task from the training set.

    We also compute the embedding of the questions and the demonstrations. We bin the embedding similarity of each demonstration/response pair into separate bins. Treating the no-context response as a benchmark, we record the general score improvement of the response within each embedding similarity bin compared to the raw benchmark.

\subsection{Context Relevancy and Performance improvement}

\begin{figure}
        \centering
        
        \begin{subfigure}[b]{0.48\textwidth}  
            \centering
            \includegraphics[width=\textwidth]{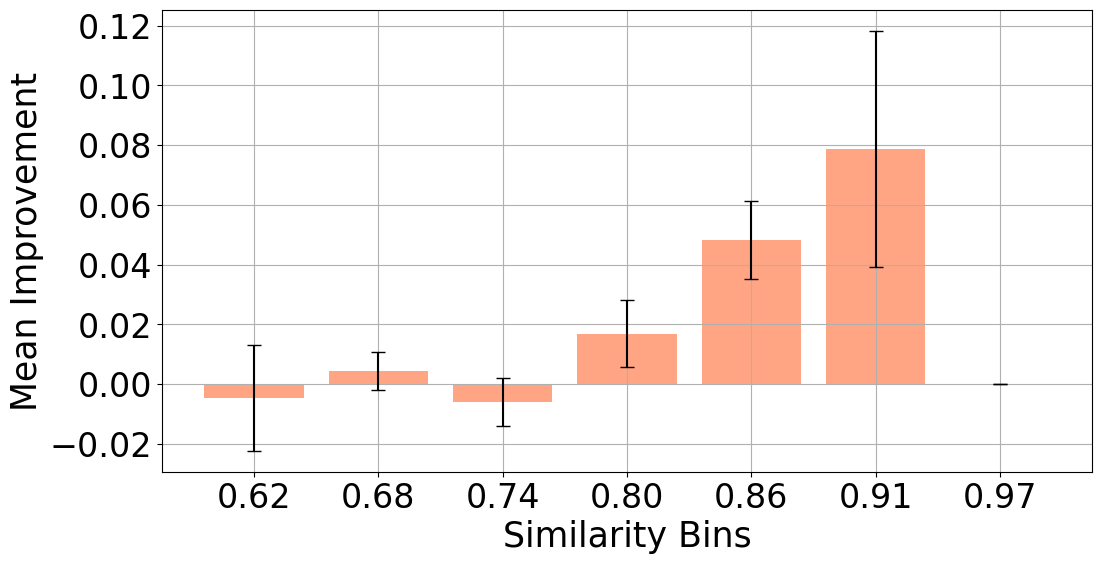}
            \caption{The MetaICL dataset contains close-form questions. The last bin is insignificant as it contains only 7 samples of data. The relationship between similarity and score improvement is positively correlated.}
            \label{fig:figure2}
        \end{subfigure}
        \hfill  
        \begin{subfigure}[b]{0.48\textwidth}  
            \centering
            \includegraphics[width=\textwidth]{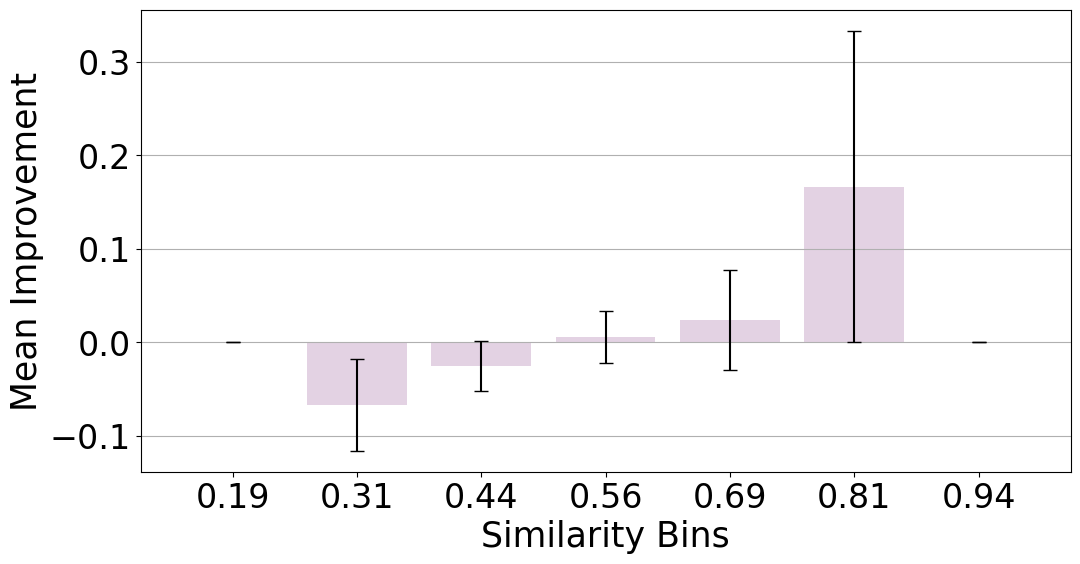}
            \caption{The NephSAP dataset contains close-form questions. The first bin and the last bin are insignificant as they contain only 1 sample each. The relationship between similarity and score improvement is positively correlated.}
            \label{fig:figure3}
        \end{subfigure}
        
        \vspace{0.5cm}  
        \begin{subfigure}[b]{0.48\textwidth}  
            \centering
            \includegraphics[width=\textwidth]{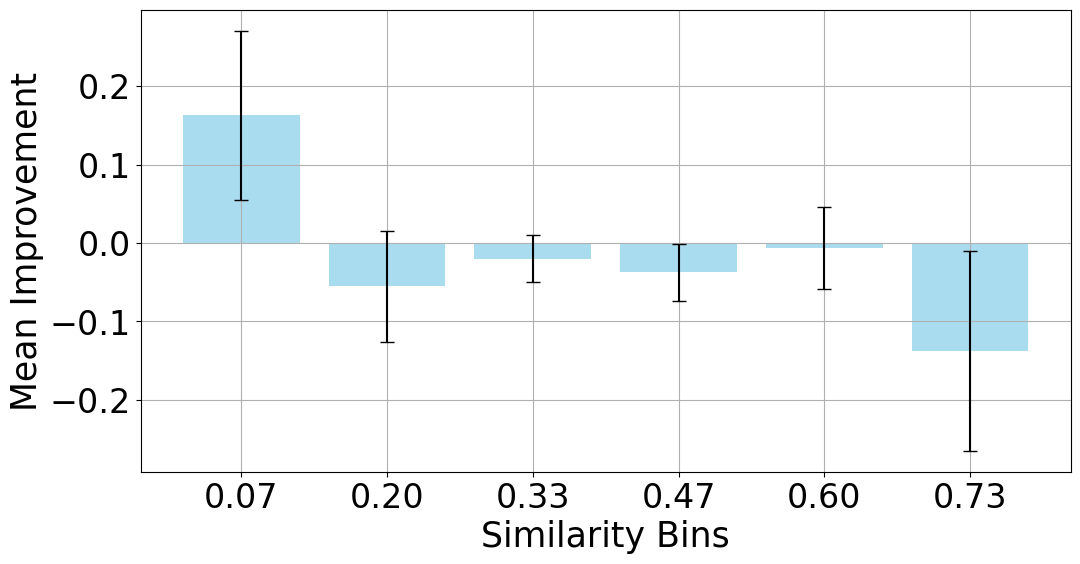}
            \caption{The dataset contains open-form physics questions. The relationship between similarity and score improvement is anti-correlated.}
            \label{fig:figure1}
        \end{subfigure}
        
        \caption{Comparison of results for different datasets. (a) Results for the MetalCL dataset. (b) Results for the NephSAP dataset. (c) Results for the Open dataset.}
        \label{fig:mergedfigures}
    \end{figure}

    In Fig. \ref{fig:mergedfigures}, we show the score improvement as a result of different contexts, using the no-context answer as the baseline.

    Note how context similarity is positively correlated with the mean score improvement in both of the closed-question datasets (MetalICL and NephSAP). This result is consistent with the arguments made by \citet{DBLP:journals/corr/abs-2101-06804} and \citet{rubin-etal-2022-learning}. Note also that in both closed-question datasets, the context with the lowest levels of similarity scores has a tendency to have a negative mean improvement (meaning, adding context hurts the results). As contexts with low levels of similarities are more likely to be contexts with a different subject or task, this result is consistent with the findings in \citep{DBLP:journals/corr/abs-2101-06804}, where irrelevant demonstrations can hurt the performance of LLM.

    This contrasts the results for the closed-form questions, as depicted in Fig. \ref{fig:mergedfigures}, panel C. Our open-form question results display a negative correlation between context similarity and mean improvement. The results suggest that, in this case, context with a lower level of similarity can be more helpful in improving the quality of the response, whereas context with a higher level of similarity can hurt the quality of the response.

\section{Discussion}

\subsection{Impact of our work and the future directions}

\begin{figure}
        \centering
        \includegraphics[width=1\textwidth]{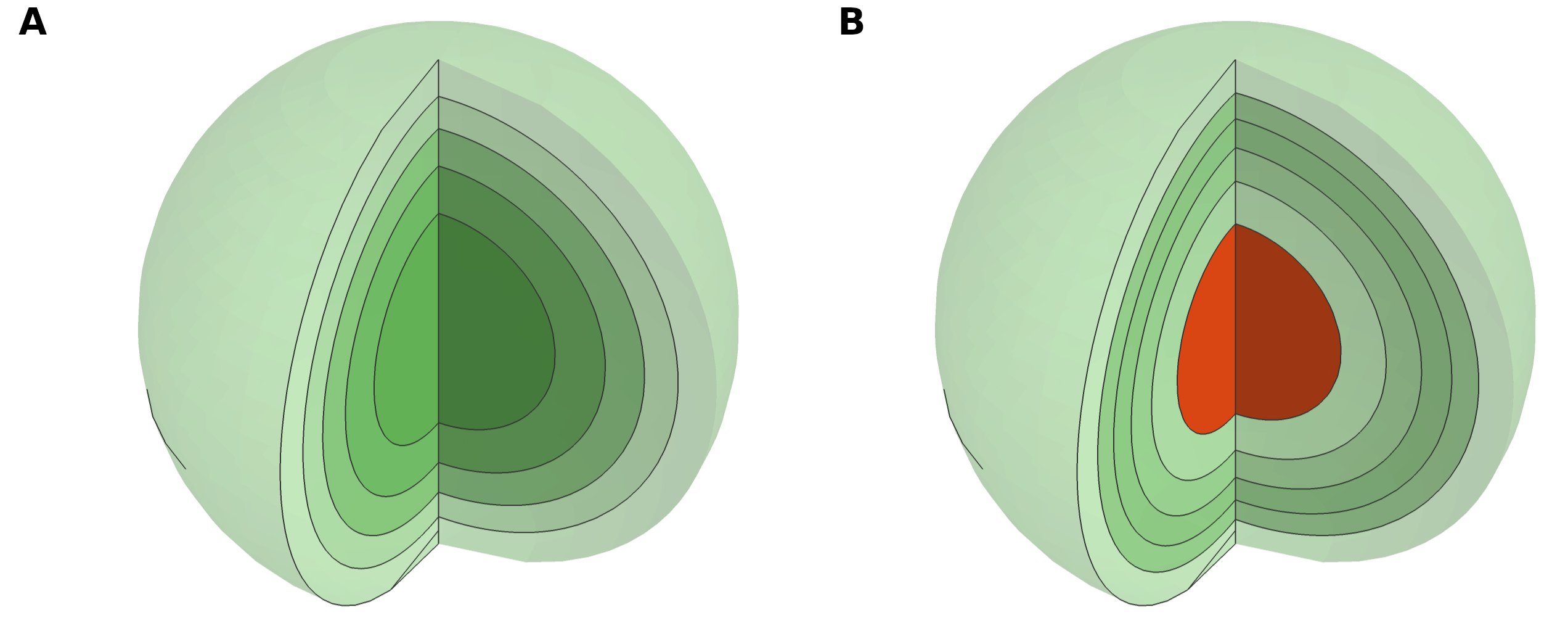}
        \caption{(A) A typical hypershpere, from which we sample documents in RAG applications. (B) An alternative approach, where we either exclude or at least diminish the impact of contexts that are too close to the point representing the querry.}
        \label{fig:donut_rag}
    \end{figure}

    Our results have suggested a significant difference between open-form question evaluation and close-form question evaluation, as the relationship between context-similarity and performance improvement is completely reversed in those two cases. The implications of this result are twofold. First, the difference between open-question evaluation and close-question evaluation invokes a new discussion on their different applicability in the context of in-context learning. Second, those mixed results suggest that similarity score might not be the best indicator for context selection in in-context learning, especially in cases that involve open-form questions. This has profound implications, especially in the context of Retrieval Augmented Generation (RAG) applications.
    For example, instead of selecting all points that lie in the vicinity of a certain point in the embedding space representing a query (cf. Fig.\ref{fig:donut_rag}, panel A), a better choice could be to either exclude or at least diminish the impact of contexts that are too close to that point (cf. Fig.\ref{fig:donut_rag}, panel B). This would lead to more interesting topologies. Instead of sampling the context from a hypersphere, we could sample from shells of various thicknesses.
    

\subsection{How should we evaluate in-context-learning? Open vs Close}

    The different behaviors exhibited in open-form question evaluation and closed-form question evaluation stem from a different treatment of context in those two cases. We provide a hypothetical interpretation of that mechanism. 
    In closed-form multiple-choice questions, the evaluated language model is treated as a classification model. A relevant demonstration provided as a context can improve the LLM's performance by aligning it with the correct choice. 
    In open-form questions, the evaluated language model is treated as a generative model, and the response is open-form. Instead of being either correct or incorrect, an open-form response can be anywhere in between. A relevant context provides alignment with one way of approaching the question, but it can also introduce bias, leading to performance degradation instead of improvement.
    
\subsection{How should we select context with respect to RAG}

    The difference between the relationship between context relevancy and performance in open-form and closed-form questions suggests that the RAG is highly application-dependent. For example, the strategy for context retrieval for open-form applications should be different from the strategy used in closed-form applications. It is also important to be mindful when evaluating RAG, as common closed-form benchmarks might not be good indicators of RAG's performance in open-form applications.
    When designing an RAG, especially in open-form applications, it is important to include some other factors than pure embedding distance or relevancy. Sometimes including a piece of context that is not as close in embedding distance to the question might be helpful as it does not reinforce the hidden bias inside the question. 
    
\bibliographystyle{plainnat}
\bibliography{bibliography}

\section*{Data and Code Availability}

    Data and code can be found in the following GitHub repository: \href{https://github.com/mikelixiang88/context-matters.git}{https://github.com/mikelixiang88/context-matters.git}

\section*{Acknowledgements}

    We would like to take this opportunity to thank Professor Stephan Haas for helpful discussions at the early stage of this project and Anurag Maravi for his engagement during the preliminary stage of the work.

\section*{Author Contributions}

    X.L., H.T., and M.A. contributed to the conceptual design,
    X.L. and H.T. developed the Python code and conducted the experiments,
    X.L., H.T., S.C., and M.A. analyzed and interpreted the results.
    All authors equally contributed to the creation of the novel dataset,
    M.A. provided supervision and proposed the experiment measuring the impact of the context.
    All the authors contributed to writing the article.
    
\section*{Competing Interests}

    The authors declare no competing interests.

\clearpage
\appendix

\section*{\Large Supplementary Material}

\section{Sample question}
\subsection{Sample Question for Open Dataset}
\textbf{Question:} Given the wavelength of an electron is \(0.364 \cdot 10^{-9} \, \text{m}\), calculate the speed of the electron.

\textbf{Ground Truth for Grading:} \\
\(\lambda = 0.364 \times 10^{-9} \, \text{m}\) \\
Mass of electron, \( m = 9.1 \times 10^{-31} \, \text{kg} \) \\
Planck's Constant, \( h = 6.62607015 \times 10^{-34} \, \text{Js} \) \\
The de Broglie wavelength is given by \( \lambda = \frac{h}{mv} \) \\
Velocity of the electron, \( v = 2 \times 10^6 \, \text{ms}^{-1} \)

\textbf{Relevant Context}\\
The De Broglie states that $\lambda = \frac{h}{{mv}}$. The mass of an electron is about $9.109 \cdot 10^{-31} kg$

\textbf{Vague Context}\\
Wave-particle duality is the concept in quantum mechanics that quantum entities exhibit particle or wave properties according to the experimental circumstances.

\textbf{Irrelevant Context}\\
Quantum physics is the study of matter and energy at the most fundamental level. At very small scale, classical theories may not be applicable any more. That is where quantum theories come into play.

\subsection{Sample question for MetaICL dataset}
\textbf{Test Input:} \\
Bird feet can also vary greatly among different birds. Some birds, such as gulls and terns and other waterfowl, have webbed feet used for swimming or floating (Figure below). Other birds, such as herons, gallinules, and rails, have four long spreading toes, which are adapted for walking delicately in the wetlands (Figure below). You can predict how the beaks and feet of birds will look depending on where they live and what type of food they eat. Flightless birds also have long legs that are adapted for running. Flightless birds include the ostrich and kiwi. Some birds, such as gulls and terns and other waterfowl, have what type of feet used for swimming or floating?

\bigskip

\textbf{Test Output:} \\
webbed

\bigskip

\textbf{Test Options:}
\begin{enumerate}
    \item lobed
    \item quad toed
    \item bipedal
    \item webbed
\end{enumerate}

For our task selections from the MetaICL dataset, please visit our GitHub repository, where the task category selections and code are presented.

\subsection{Sample question for NephSAP dataset}

A 54-year-old man with ESRD is admitted for management of presumed catheter–related bacteremia. He had no pre–ESRD nephrology care and recently started maintenance hemodialysis on an urgent basis for symptomatic uremia. Two days ago, he developed acute onset of fever to 40 C, chills, and rigors during dialysis. After obtaining blood cultures, he received intravenous vancomycin and ceftazidime at the outpatient dialysis center before admission. His tunneled dialysis catheter is removed because of erythema and purulent exudate at the exit site. His fever is now resolved. He does not have back pain, cough, dental complaints, or rash. On physical examination, he is now afebrile. The BP is 136/88 mmHg. There is a 1-cm area of erythema at the tunneled dialysis catheter removal site. He has no stigmata of emboli. He has no obvious dental issues on gross inspection of the oropharynx. The cardiac examination shows a grade 2 early peaking systolic crescendo-decrescendo murmur along the right sternal border without radiation. The remainder of the examination is unremarkable. The blood cultures drawn before implementation of antibiotic therapy grew \textit{S. aureus} within 12 hours. The results of sensitivity testing and repeat blood cultures are pending. Laboratory studies show sodium 138 mEq/L, potassium 3.8 mEq/L, chloride 102, total CO2 25 mmol/L, BUN 45 mg/dl, and creatinine 6.4 mg/dl.

In addition to continued antibiotic therapy, which ONE of the following is the next BEST step in management?

\begin{itemize}
    \item[A.] Arrange for placement of a tunneled dialysis catheter today
    \item[B.] Transesophageal echocardiography
    \item[C.] Magnetic resonance imaging of the spine
    \item[D.] Bone scintigraphy (bone scan)
    \item[E.] Examination for an occult dental source of bacteremia
\end{itemize}

\section{Context types}

\section{Generated Answer}
\subsection{Sample Response for Question in A.1}

\textbf{Response with No Context:} \\
To calculate the speed of the electron, we can utilize de Broglie's equation that relates the momentum of a particle to its wavelength. Here is the formula:
\[
\lambda = \frac{h}{p}
\]
Where:
\begin{itemize}
    \item \( \lambda \) is the wavelength
    \item \( h \) is Planck's constant
    \item \( p \) is the momentum.
\end{itemize}
If we rearrange the terms to express momentum:
\[
p = \frac{h}{\lambda}
\]
The momentum of a particle can also be described as its mass times its velocity:
\[
p = m \cdot v
\]
We know that the mass of an electron (\( m \)) is \( 9.11 \times 10^{-31} \) kg. And Planck's constant (\( h \)) is \( 6.626 \times 10^{-34} \) Js. \\
So the velocity (\( v \)) is given as:
\[
v = \frac{h}{m\lambda}
\]
Substituting the known values, we get:
\[
v = \frac{6.626 \times 10^{-34}}{9.11 \times 10^{-31} \cdot 0.364 \cdot 10^{-9}}
\]
After solving this, the velocity of the electron comes out to be approximately \( 7.24 \times 10^5 \) m/s.

\bigskip

\textbf{Response with Irrelevant Context:} \\
The speed of an electron given its wavelength can be calculated using the de Broglie's equation:
\[
\lambda = \frac{h}{mv}
\]
where:
\begin{itemize}
    \item \( \lambda \) is the wavelength,
    \item \( h \) is Planck's constant, which is roughly equals to \( 6.626 \times 10^{-34} \, \text{Js} \),
    \item \( m \) is the mass of the electron, which is roughly equals to \( 9.109 \times 10^{-31} \, \text{kg} \),
    \item \( v \) is the velocity of the electron.
\end{itemize}
We can rearrange the equation to solve for the velocity:
\[
v = \frac{h}{m\lambda}
\]
Substitute the given values into the equation:
\[
v = \frac{6.626 \times 10^{-34} \, \text{Js}}{9.109 \times 10^{-31} \, \text{kg} \times 0.364 \times 10^{-9} \, \text{m}}
\]
and calculate the velocity.

\bigskip

\textbf{Response with Vague Context:} \\
The wavelength of a matter wave, such as an electron, can be found using De Broglie's equation, which is given by:
\[
\lambda = \frac{h}{p}
\]
where:
\begin{itemize}
    \item \( \lambda \) is the wavelength
    \item \( h \) is Planck's constant (\( 6.626 \times 10^{-34} \, \text{Js} \))
    \item \( p \) is the momentum of the particle, which can be expressed as \( p = mv \)
\end{itemize}
Since we want to find the speed of the electron (\( v \)), we can rearrange this equation in terms of \( v \):
\[
v = \frac{h}{m\lambda}
\]
Given that \( \lambda = 0.364 \times 10^{-9} \, \text{m} \) and the mass of an electron is \( m = 9.109 \times 10^{-31} \, \text{kg} \), we can substitute these values into the equation:
\[
v = \frac{6.626 \times 10^{-34}}{9.109 \times 10^{-31} \times 0.364 \times 10^{-9}}
\]
Evaluating this will give the speed of the electron in m/s.

\bigskip

\textbf{Response with Relevant Context:} \\
The De Broglie equation relates the wavelength of a particle to its momentum. According to De Broglie's principle, every object in motion has a wave character. The De Broglie wavelength is given by the formula:
\[
\lambda = \frac{h}{mv}
\]
where:
\begin{itemize}
    \item \( \lambda \) is the wavelength,
    \item \( h \) is Planck's constant (\( 6.626 \times 10^{-34} \, \text{m}^2 \cdot \text{kg} / \text{s} \)),
    \item \( m \) is the mass of the particle,
    \item \( v \) is the velocity (or speed) of the particle.
\end{itemize}
We are given that \( \lambda = 0.364 \times 10^{-9} \, \text{m} \), and the mass \( m \) of an electron is \( 9.11 \times 10^{-31} \, \text{kg} \). We can find the speed \( v \) by rearranging the above formula:
\[
v = \frac{h}{m\lambda}
\]
Substitute the known values into the rearranged equation to find \( v \), the speed of the electron:
\[
v = \frac{6.626 \times 10^{-34} \, \text{m}^2 \cdot \text{kg} / \text{s} }{(9.11 \times 10^{-31} \, \text{kg}) (0.364 \times 10^{-9} \, \text{m})}
\]
Solving the above expression will give us the speed of the electron.
\section{Grading Interface}
To better evaluate the open response to our physics questions, we modified the potato annotation system \citep{pei2022potato} and applied it as our evaluation system. Our evaluation system not only allows users to select numeric grades for each response but also enables the user to highlight parts of the response, apply labels, and write descriptions to justify their grading. In addition, the system randomly shuffles the order of the responses for each grader to mitigate any potential bias in grading as a result of the ordering of responses. A short video tutorial is provided at the beginning page to provide guidance and alignment in grading.

\begin{figure}[b!]
    \centering
    \includegraphics[width=1\textwidth]{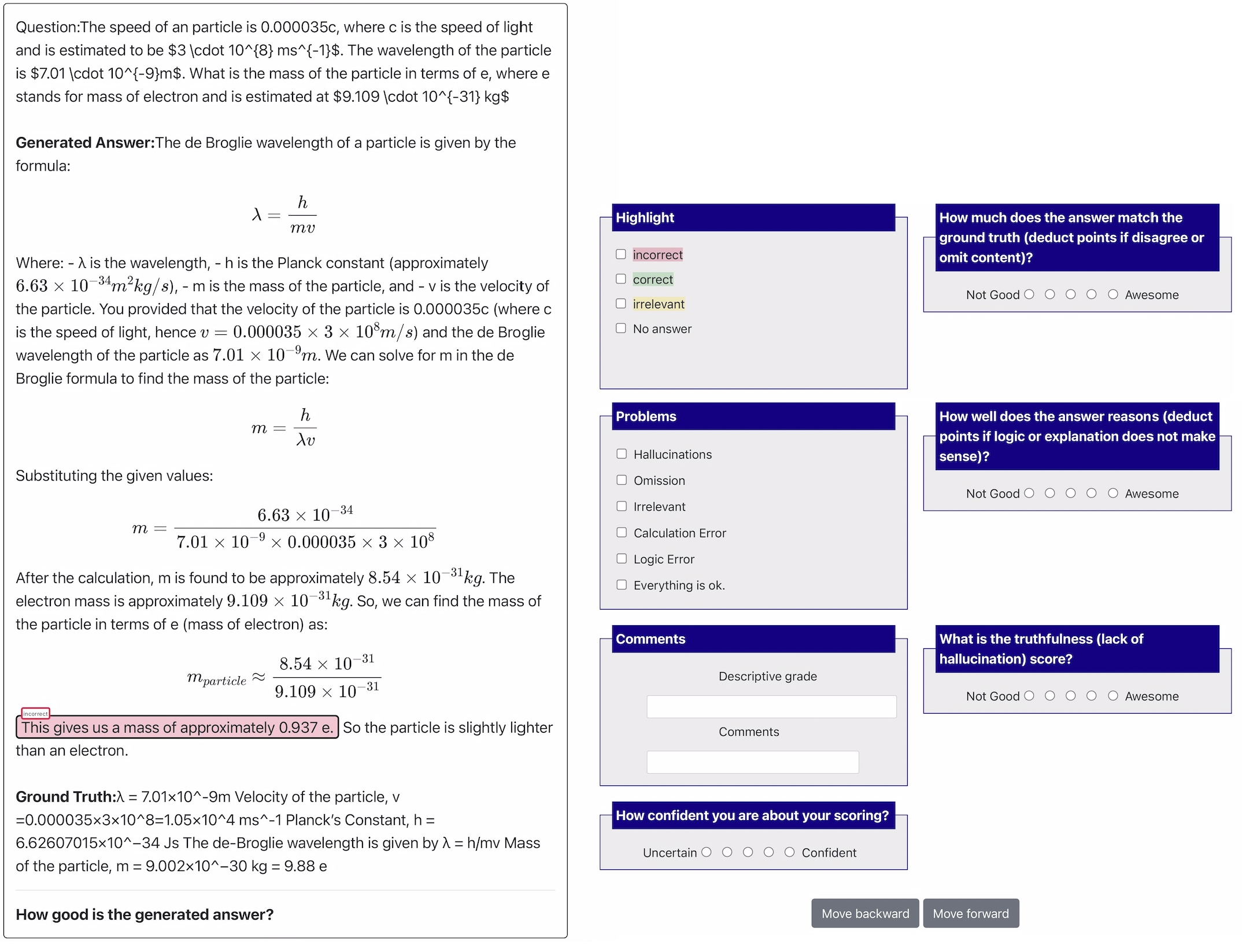}
    \caption{The potato grading interface used in evaluation} 
    \label{fig:potato_interface}
\end{figure}

A screenshot of the interface of the evaluation system is shown in Fig.~\ref{fig:potato_interface}. The system is also accessible via the link: \href{http://quantumgpt.science:8080/?PROLIFIC_PID=testuser}{\url{http://quantumgpt.science:8080/?PROLIFIC_PID=testuser}}.

\section{Sanity Check}

\begin{figure}[b!]
        \centering
        \includegraphics[width=0.5\textwidth]{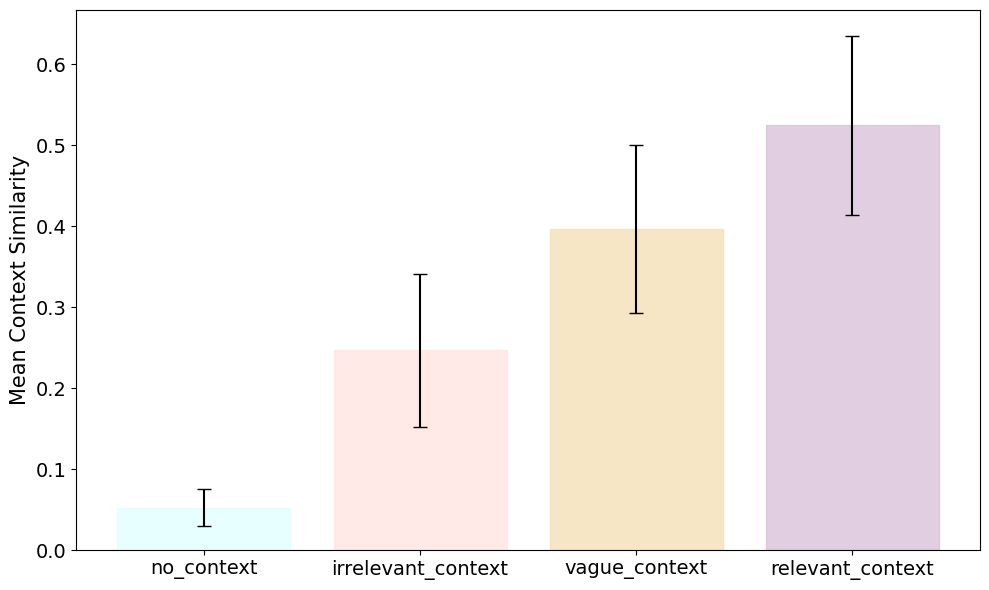}
        \caption{Mean context similarity by context type.} 
        \label{fig:mean_context_similarity_by_context_type}
    \end{figure}

    To check whether our context relevancy is well defined, we compute the embedding of the questions and their respective contexts for both our open-form question dataset and the two closed-form question datasets we use. We then calculate the cosine distance between the embedding of each question and the different contexts associated with them. We show the results for the open question dataset in Fig.~\ref{fig:mean_context_similarity_by_context_type}.

\begin{figure}[b!]
        \centering
        \includegraphics[width=1\textwidth]{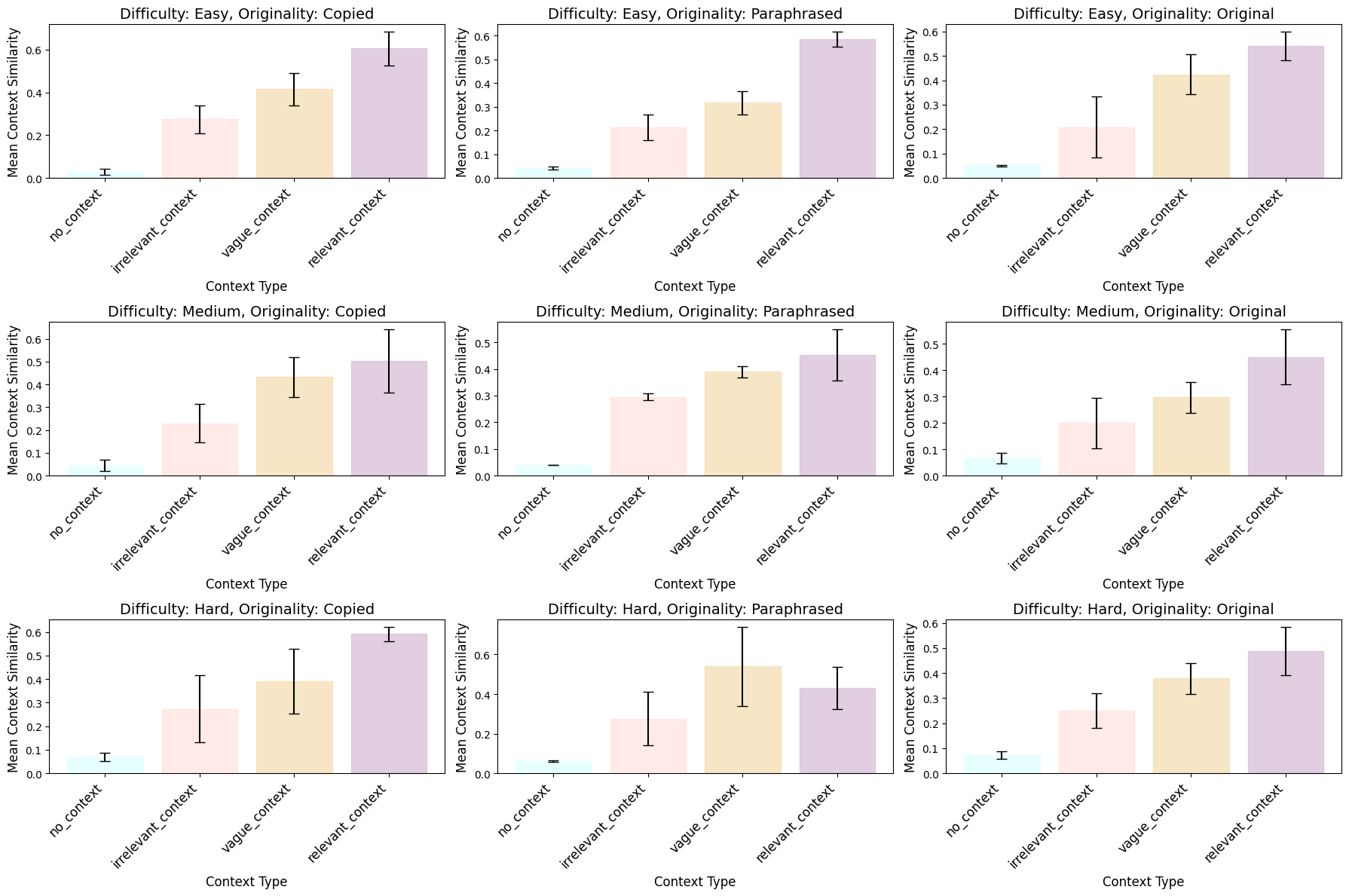}
        \caption{Mean context similarity by context type across different originality and difficulty.}
        \label{fig:mean_context_similarity_by_context_type_break_down}
    \end{figure}

    We computed the embedding of each question and each context using OpenAI's ``text-embedding-3-large'' model. For the no-context part, we used a space as a placeholder instead of an empty string. 
    As expected, the results show that more relevant contexts, as perceived by us when designing the dataset, receive a higher mean similarity score with their respective questions. Different question types can result in a large standard deviation in similarity scores in different contexts.
    We show the details breakdown of those results in Fig.~\ref{fig:mean_context_similarity_by_context_type_break_down}.

    All question types except hard paraphrased questions display the same trend, confirming the relationship between context types and embedding similarities. 

\begin{figure}[h]
        \centering
        \includegraphics[width=1\textwidth]{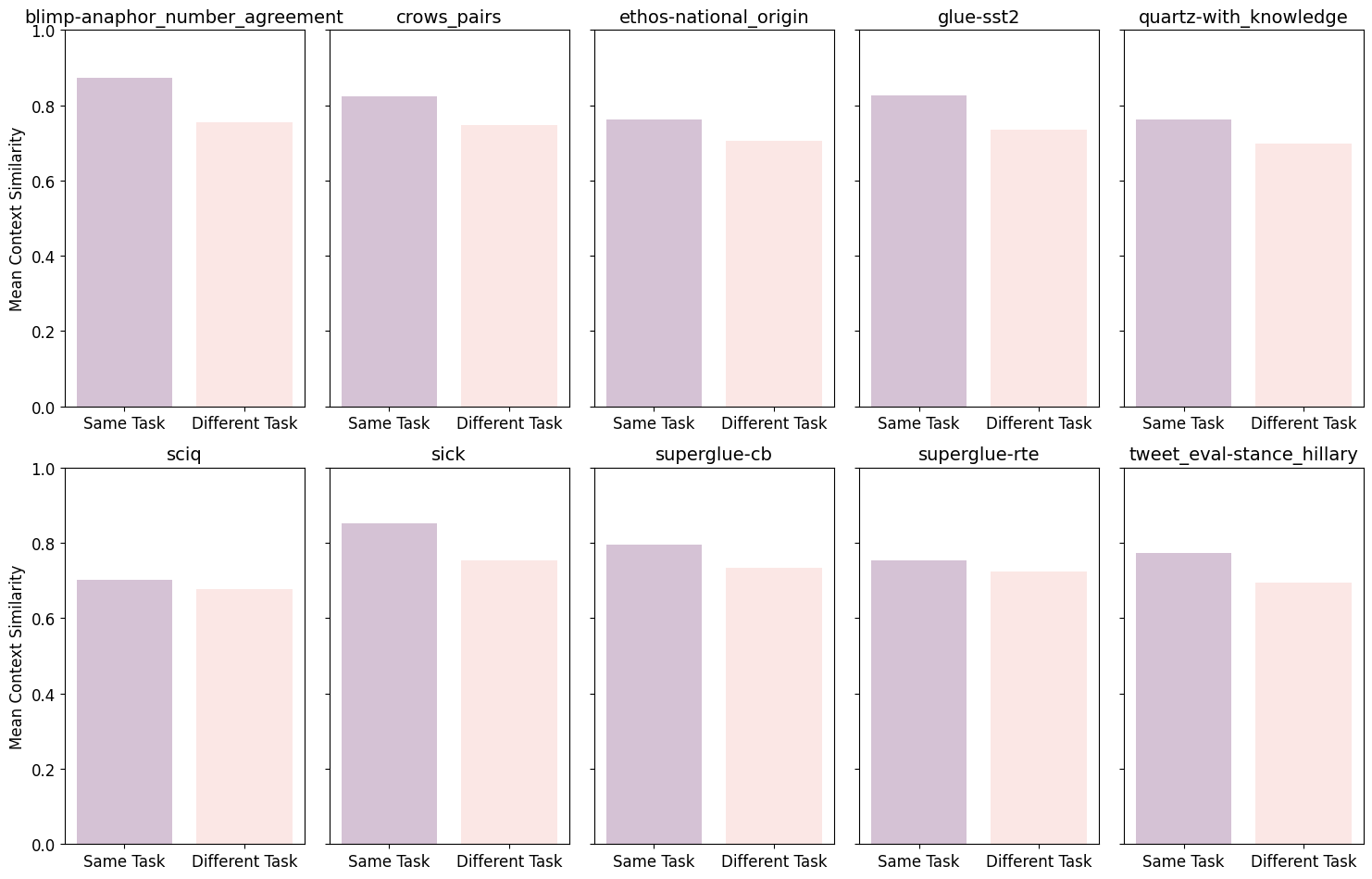}
        \caption{Mean context similarity by demonstration type across different MetaICL tasks.} 
        \label{fig:mean_context_similarity_MetalICL}
    \end{figure}

    For the closed datasets, the similarity score between context and question is shown in Table \ref{tab:similarity}. For both datasets, the same task/subject demonstrations possess a higher mean similarity score than the different task/subject demonstrations. To further verify this relationship, we have also plotted the similarity score of the same task demonstrations and different task demonstrations for each task in the MetaICL dataset in Fig. \ref{fig:mean_context_similarity_MetalICL}. The results confirm that the same task demonstration displays higher mean similarity than the different task demonstration in every task in the dataset.

    \renewcommand{\arraystretch}{1.5}
    \begin{table}[h]
    \centering
    \caption{Mean context similarity for closed datasets}
    \label{tab:similarity}
    \begin{tabular}{c|c|c}
    \hline\hline
        \textbf{Dataset} & \textbf{Average Different Task Similarity} & \textbf{Average Same Task Similarity} \\ \hline
        MetaICL           & 0.719                                     & 0.787                                 \\ \hline
        NephSAP           & 0.443                                     & 0.557                                 \\\hline\hline
    \end{tabular}
    \end{table}

\end{document}